\documentclass[runningheads]{LLNCS/llncs}

\usepackage{graphicx}
\usepackage[ruled,vlined]{algorithm2e}
\usepackage{comment}
\usepackage{float}
\usepackage{hyperref}
\usepackage{amssymb}

\def\RR {\mathbin{{\rm I}\mkern - 4mu{\rm R}}}

\begin{document}
\title{Effective Universal Unrestricted Adversarial Attacks using a MOE Approach
}
\titlerunning{Effective Universal Unrestricted Attacks }
%
\author{Alina Elena Baia\orcidID{0000-0001-5553-776X} \and
Gabriele Di Bari\orcidID{0000-0002-8341-8925} \and
Valentina Poggioni\orcidID{0000-0002-7691-7478}}

\authorrunning{A.E. Baia et al.}
%
\institute{University of Perugia, Perugia, Italy \\ \email{baia\_alinna@yahoo.com}, 
\email{dbgabri@gmail.com},  
\email{valentina.poggioni@unipg.it}}
\maketitle             
%
\begin{abstract}
Recent studies have shown that Deep Leaning models are susceptible to adversarial examples, which are data, in general images, intentionally modified  
to fool a machine learning classifier. In this paper, we present a multi-objective nested evolutionary algorithm to generate universal unrestricted 
adversarial examples in a black-box scenario.
The unrestricted attacks are performed through the application of well-known image filters that are available in several image processing libraries, modern cameras, and mobile applications. The multi-objective optimization takes into account not only the attack success rate but also the detection rate.
Experimental results showed that this approach is able to create a sequence of filters capable of generating very effective and undetectable attacks.

\keywords{Universal Adversarial Attacks \and Evolutionary Algorithms \and Multi-Objective Optimization \and Deep Learning}
\end{abstract}
\section{Introduction}

Deep learning (DL) has witnessed a significant progress over the last decade and it has been effectively applied to a variety of applications in different machine learning domains achieving state-of-the-art performance. 
The great success of DL models, both in academia and industry, made them object of attacks. People started investigating the vulnerability and security aspects of such models since attacks pose significant risks and challenges for real-world security-sensitive systems such as medical diagnosis, voice controllable systems, and autonomous driving. Recent studies have shown that Deep Neural Networks, despite their superior performance, are remarkably vulnerable to adversarial attacks, creating severe security issues at the time of deployment of such systems.

The attacking techniques, at the highest level, are classified in {\it per-instance attacks} and {\it universal attacks}. In the first class we can find all those systems that generate a different perturbation for each image; in that case a separate optimization process has to run for each image in order to find the corresponding adversarial image \cite{Deepfool,Edgefool,Colorfool,Carlini2017TowardsET,Goodfellow2015ExplainingAH,Kurakin2017AdversarialEI,Szegedy2014IntriguingPO}. On the other hand, in the second class we can find all those systems able to find a unique universal perturbation that, when applied to 'any' image, can fool the classification system; these systems are called universal because they are essentially image-agnostic \cite{MoosaviDezfooli2017UniversalAP,hayes2018learning,mopuri2018generalizable,reddy2018ask}.

Moreover, the adversarial attacks follow the typical classification used in security sectors that distinguish white-box attacks, i.e. attacks having access to the target network model and underlying training policy, from black-box ones in which the parameters and the underlying architecture are unknown to the attacker.

If we consider the type of the applied perturbations, the attacks can be classified as restricted or unrestricted. In the restricted case, the modifications 
applied to the original image are usually small
and bounded by a $L_p$-norm distance measure, forcing the adversarial image $x^{*}$ to be as close as possible to the original one. On the contrary, unrestricted attacks use large perturbations without $L_p$-bounded constraints that manipulate the image in order to create photo-realistic adversarial examples. In this case the objective is not to limit the modifications on pixels but limit the human perception that a modification has been applied \cite{Colorfool}.

From the point of view of the expected results, the attacks are further distinguished in \textit{untargeted}, when the aim is to simply generate a misclassification, and \textit{targeted}, when the misclassification is driven towards a specified target class. 

In this scenario, the robustness of DNN against adversarial examples has gained significant attention in the last few years, and several approaches and systems able to detect adversarial attacks have been proposed and developed. Some of them follow the {\it adversarial training } approach increasing the network robustness by means of adversarial examples in the training process \cite{Szegedy2014IntriguingPO,Deepfool,kurakin2016adversarial,tramer2018ensemble}, while others propose ad-hoc trainable techniques like distillation \cite{Papernot2016DistillationAA}, perturbation rectifying \cite{Akhtar2018DefenseAU}, feature squeezing \cite{Xu2018FeatureSD}
input reconstructions \cite{Qin2020DetectingAD}, input transformation \cite{JPEG}, input denoising \cite{Meng2017MagNetAT}, and many more. Although these defense mechanisms were inspired by different perspectives, they are mostly limited to safeguard against norm-bounded attacks.

 
In a black-box scenario, traditional approaches rely on gradient estimation or on training a substitute network and transfer the generated examples to the targeted model \cite{Chen2017ZOOZO,Papernot2017PracticalBA,Narodytska}, but alternative, gradient-free optimization techniques, mostly based on evolutionary algorithms, have been recently introduced \cite{GenAttack,AdversarialPSO,Vidnerov2020VulnerabilityOC,Su2019OnePA}. Among this group, techniques using Multi-Objective optimization recently reached interesting results and are quickly emerging \cite{suzuki2019adversarial,deng2019multi}.

The majority of the proposed attacks are performed and optimized to add small random perturbations to the pixel values. However, these artificial modifications are often not semantically meaningful and can create unnatural-looking images that are easily detectable. For this reason, researchers start exploring new types of threats models that can significantly change an input while maintaining the semantics.

Such methods require either access to the targeted network architecture \cite{ACE} or 
additional resources like pretrained networks to perform image segmentation \cite{Colorfool}, colorization and style transfer \cite{Bhattad2020UnrestrictedAE}.
In some cases, it is necessary to train neural networks from scratch in order to find the adversarial perturbations \cite{Edgefool}. 

In our work, we decided to focus on generating non-targeted unrestricted universal adversarial attacks in a black-box scenario, 
since the limited knowledge and ability of the attacker is more similar to a real-world scenario, making the attack itself more challenging but its applicability more practical.

We propose a gradient-free method based on  nested evolutionary algorithms and multi-objective optimization that, given a set of commonly-used image filters, finds an optimal image-agnostic sequence of them that, when applied to an image is hardly detectable and causes the classifier to misclassify the image. The standard universal $L_p$-bounded attacks are transformed into universal unrestricted attacks and a multi-objective evolutionary approach is used to build a process able to optimize, at the same time, the attack success rates and the attack detection rate.  

By using well-known filters already available in several image processing libraries and in modern cameras and  widely used in social media (e.g. Instagram), we aim to reduce the awareness towards the applied modification. 

The method provides two optimization stages. The first stage utilizes a genetic algorithm in order to identify the optimal sequence of filters, whereas the second one optimizes the parameters of each selected filter.
For this stage we investigated three optimization strategies: genetic algorithm, evolutionary strategy and a random approach with tournament. 

To find a successful adversarial filter configuration, a population of candidate solutions is evolved and the solution quality is assessed by means of a fitness value. Our goal is to select filter sequences that will not alter the shape and semantics of images while maximizing the attack success rate on the target model. Moreover, since many deployed deep learning models are protected by defense methods, we want to find an attack able to bypass such mechanisms. Thus, we choose to incorporate the feedback given by the defenses methods directly into the fitness function used for the filters optimization.

Given the conflicting nature of the above-mentioned objectives and motivated by the success of multi-objective evolutionary algorithms (MOEA) in other applications \cite{MOEA_survey,MOEGAN}, we propose to model our method as a multi-objective optimization problem. We employ the non-dominated sorting genetic algorithm II (NSGA-II) for the selection process. 

The experimental results demonstrates the effectiveness of our method when tested against one of the most highly rated detection frameworks, namely Feature Squeezing \cite{Xu2018FeatureSD}. Our algorithm is able to bypass such defense in most of the cases, having a detection rate smaller than 5\% on the testing set while achieving good results in term of attack sucess rate.

\section{Related works}
Over the years, many methodologies have been proposed for generating adversarial examples in both white-box and black-box settings. 

Szegedy et al. \cite{Szegedy2014IntriguingPO} were the first to introduce the concept of adversarial examples by analyzing the properties of neural networks that make these models susceptible to adversarial attacks. The authors used box-constrained L-BFGS to calculate the perturbation needed to get the image misclassified. Based on this work, Goodfellow et at. \cite{Goodfellow2015ExplainingAH} explained that the linear part of the high-dimensional model is to blame for their sensitivity to small changes in the input. They also introduced a fast method for generating adversarial examples (FGSM). In the following years, a variety of other attack algorithms have been proposed, both in the white-box \cite{Carlini2017TowardsET,Papernot2016TheLO,Deepfool} and black-box scenario \cite{Papernot2017PracticalBA,Chen2017ZOOZO,MoosaviDezfooli2017UniversalAP}.



Most of the proposed works on adversarial examples have been focusing on finding small perturbations that can change the predictions of a classifier: in some cases it is sufficient to change just one pixel \cite{Su2019OnePA} or inject a random quasi-imperceptible-perturbation \cite{MoosaviDezfooli2017UniversalAP}. Due to the urgency of taking counter-measures, several detection and defense methods have been introduced to overcome such vulnerabilities. Therefore, nowadays, these types of adversarial images are easily detectable by applying denoising filters or by adversarial training \cite{AdvExSurvey}. For this reason, many researchers have been shifting their attention to unrestricted adversarial attacks that employ large and visible perturbations but have the advantage that the resulting images are still looking natural and non-suspicious to the human eye.

Hosseini et al. \cite{Semnatic_adv_ex} were one of the first to analyze the effectiveness of unrestricted adversarial examples on deep learning models. They proposed to randomly change the hue and saturation values of an image while maintaining the shape of the objects. The authors in \cite{Colorfool} address the limitations of \cite{Semnatic_adv_ex} which was found to produce unnatural colors by employing priors on color perception. Other works propose to craft malicious inputs by applying image-enhancement filters obtained by means of neural networks \cite{Edgefool} or via gradient descent optimization \cite{ACE}. Pretrained colorization models and texture style transfer methods have also been successfully utilized without norm constraints on the perturbations \cite{Bhattad2020UnrestrictedAE}.

Our idea is to use well known filters that are available in several libraries and mobile applications used extensively every day to enhance photos and images. The power of this application relies on the natural presence of these filters in almost all the images we can find everywhere and this essentially makes them transparent to the human perception. Clearly, the filters have to be ``gently'' applied otherwise the resulting image could become unrealistic, for example with supersaturated colors, but in general they cannot alter the image semantic. 

 




In the world of adversarial attacks, recent studies have proposed the use of evolutionary algorithms to overcome the limitations imposed by the methods relying on gradient computation/estimation as well as the long training time necessary in the generative approaches.
The main benefit is that a population-based optimization algorithm does not require gradient computation nor a differentiable objective function. Furthermore, the gradient-free nature of such methods makes the attacks more robust to gradient masking and obfuscation defenses. Several works which employ a variety of evolutionary algorithms such as Differential Evolution, Genetic Algorithm, and Particle Swarm Optimization have been presented.
All of them evolve a population of feasible solutions, according to a fitness function. They make use of similar strategies: the new candidate solutions are obtained by applying small random perturbations to the initial population and the fitness of each population individual is evaluated to find a successful example \cite{Su2019OnePA,GenAttack,Vidnerov2020VulnerabilityOC,AdversarialPSO}. Moreover, two very recent papers introduced the idea to use 
multi-objective evolutionary algorithm to produce image perturbations
trying, at the same time, to maximize the attack success rate and 
minimize the perturbation size \cite{deng2019multi,suzuki2019adversarial}.
Differently from them, we propose to include in the optimization process, alongside the maximization of the attack rate,  the minimization of the detection rate of defence methods in order to produce attacks that will be intrinsically successful. 
To the best of our knowledge there are no other works that take into consideration the detection rate while crafting the attack. 





\section{Detection method: Feature Squeezing}
Ever since the adversarial attacks were fist introduced, there has been a continuous battle between attacks and defenses. 
Numerous countermeasures 
have been proposed to mitigate the threat of adversarial attacks. 
These include adversarial training \cite{Goodfellow2015ExplainingAH} 
, defensive distillation \cite{Papernot2016DistillationAA},  detection methods based on requantization and median filtering
\cite{Xu2018FeatureSD}, input reconstructions \cite{Qin2020DetectingAD},  
 input transformation \cite{JPEG}
, input denoising \cite{Meng2017MagNetAT}, and many more.

We choose Feature Squeezing \cite{Xu2018FeatureSD} as a detection strategy since is one of the most popular techniques that achieves high detection rates against state-of-the-art attacks. 

This method is based on the observation that often the space of feature vectors of images is unnecessarily large which gives plenty of manipulation possibilities for generating  adversarial examples. The authors proposed to squeeze out unnecessary input features in order to reduce the search space accessible to an adversary 
by means of two feature squeezing methods: color bit depth reduction  of each pixel and spatial smoothing (local and non-local smoothing). 
Therefore, it is possible to tag an image as legitimate or as adversarial  by comparing  the 
model's prediction on the original image with its prediction on the squeezed image.
If the difference between the prediction vectors of the original sample and the squeezed sample exceeds a certain threshold than the input is identified to be adversarial. Given that the selection of an optimal threshold value in not a trivial task and requires a training phase, for the experiments in this work we refer to the thresholds reported by the authors in \cite{Xu2018FeatureSD}.

The authors also show that adversarial examples from eleven state-of-the-art attacks can be successfully detected by combining multiple squeezing defenses into a joint detection framework. 


\section{Image Filters}

We employed Instagram inspired image filters to perform the attacks. The filters were implemented using Python3 and the Pillow, OpenCV and Numpy libraries. We chose five of the most popular Instagram filters, specifically Clarendon, Juno, Reyes, Gingham and Lark. Each filter has distinct characteristics and effects given by different level of contrast, saturation, brightness, shadows, etc.:
\begin{itemize}
    \item  Clarendon adds light to bright areas  darkness to dark areas, slightly increasing saturation and contrast while keeping the mid-tones rather warm and cooling down shadows and highlights.
    \item  Juno only manipulates the contrast and the vividness of an image by intensifying the yellows and reds making these colors pop out more than the blues.
    \item  Reyes adds a subtle old-time  look by reducing the saturation and by brightening up the photos.

    \item Gingham also gives a dusty-vintage feel to the image. It significantly lowers the highlights and the saturation. To complete the look it applies a white soft lens effect which draws the attention to the center of the images and creates an elegant and dreamy atmosphere.
    
    \item  Lark increases the exposure making the photo brighter and  reduces the vibrance. Moreover, it also accentuates blues and greens while desaturating the reds.
\end{itemize}
For each filter there are two parameters that the evolutionary algorithm has to tune:  {\it intensity} $\alpha$ and {\it strength} $s$.

The $\alpha$ manipulates the intensity of the filter, i.e. for Clarendon $\alpha$ determines how much the dark areas are light, while for Juno  the quantity of yellows and reds that has to be applied, whereas for Gingham that parameter indicates the intensity of the vintage effect. Following the same logic, for Lark $\alpha$  controls the increment and decrement of the photo exposure.
For the sake of clarity, in Table \ref{filter_effects} the effects of the $\alpha$ parameter for each filter is reported.

Regarding the {\it strength}, it is the parameter of the convex interpolation among the original image $x$ and the manipulated image  $x^*$, which is calculated as follows:
\begin{equation}
strength(x,x^*,s) = (1.0-s) \cdot x + s \cdot x^*
\end{equation}
thus, if $s=0$ the output image of the filter is the original image, while with $s=1$ the filter returns the manipulated image $x^*$.


\begin{table}[h]
\centering
\caption{Effects of filters with different $\alpha$ values }
\begin{tabular}{|c|c|c|c|c|c|}
\hline

&Clarendon & Juno & Reyes & Gingham & Lark\\
\hline
Original & \includegraphics[scale=1]{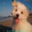} & \includegraphics[scale=1]{dog.png}& \includegraphics[scale=1]{dog.png}&\includegraphics[scale=1]{dog.png} & \includegraphics[scale=1]{dog.png}\\

\hline
$\alpha = 0.5 $ & \includegraphics[scale=1]{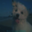}& \includegraphics[scale=1]{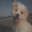}& \includegraphics[scale=1]{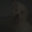}&\includegraphics[scale=1]{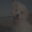} &\includegraphics[scale=1]{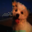} \\
\hline
$\alpha = 0.65 $ & \includegraphics[scale=1]{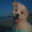}&\includegraphics[scale=1]{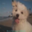} &\includegraphics[scale=1]{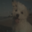} &\includegraphics[scale=1]{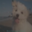} & \includegraphics[scale=1]{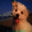}\\
\hline
$\alpha = 0.8 $ & \includegraphics[scale=1]{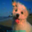}& \includegraphics[scale=1]{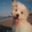}&\includegraphics[scale=1]{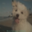} &\includegraphics[scale=1]{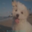} & \includegraphics[scale=1]{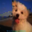}\\
\hline
$\alpha = 1.0 $ & \includegraphics[scale=1]{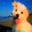}& \includegraphics[scale=1]{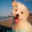}&\includegraphics[scale=1]{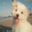} & \includegraphics[scale=1]{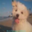}&\includegraphics[scale=1]{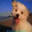} \\
\hline
$\alpha = 1.3 $ &\includegraphics[scale=1]{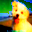} & \includegraphics[scale=1]{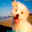}& \includegraphics[scale=1]{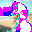}& \includegraphics[scale=1]{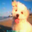}& \includegraphics[scale=1]{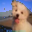}\\
\hline
$\alpha = 1.5 $ & \includegraphics[scale=1]{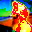}& \includegraphics[scale=1]{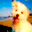}& \includegraphics[scale=1]{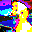}&\includegraphics[scale=1]{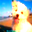} &\includegraphics[scale=1]{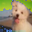} \\
\hline
\end{tabular}
\label{filter_effects}
\end{table}

\section{Problem Formulation and Backgrounds}

Given an input image $x\in X \subset \mathbb{R}^d$ and its corresponding label $y$, let $F$ be a neural network classifier that (correctly) predicts  the class label for the input image $x: F(x) = y$. An adversarial attacks attempts to modify the input image $x$ adding a perturbation $\delta$ into an adversarial image $x^{*}=x+\delta$ such that the classifier was mislead into making a wrong prediction, i.e. $F(x^{*}) \neq F(x)$. In general the objective is to find the smallest perturbation $\delta \in \mathbb{R}^d$ able to cause the misclassification and this is 
obtained limiting the perturbation size, $||\delta||_p\leq \epsilon$, where $||\cdot||_p$ denotes the $L_p$ norm defined as 
\begin{equation}
||x||_p = \sqrt[p]{\sum_{i=1}^n x_i^p}, \qquad x=(x_1,...,x_n)
\end{equation}

In case of {\it per-image} approaches, a different $\delta$ is found for each image and it is necessary to run the training process for each attack. On the other hand, in case of {\it universal} approaches the objective is to find {\it only one} such $\delta$ able to fool $F$ for {\it almost all} the data points available in $X$, that is 
\begin{equation}
F(x+\delta)\neq F(x),\qquad \textrm{for almost all }  x \in X    
\end{equation}

\subsection{Multi-Objective Problem}

\label{sec:background:mo}

In multiobjective optimization, the aim is to solve problems of the 
type\footnote{Without loss of generality, we will assume only minimization problems.}:

\begin{equation}
\textnormal{minimize } \vec{f}(\vec{x}):=\left[ f_{1}(\vec{x}), f_{2}(\vec{x}), \ldots, f_{k}(\vec{x}) \right]
\end{equation}
\noindent subject to:
\begin{equation}
g_{i}(\vec{x}) \leq 0 \ \ \ \  i=1,2,\ldots,m
\label{eq:one}
\end{equation}
\begin{equation}
h_{i}(\vec{x}) = 0 \ \ \ \  i=1,2,\ldots,p
\label{eq:two}
\end{equation}

\noindent 
where $\vec{x}=\left[ x_{1}, x_{2}, \ldots, x_{n} \right]^{T}$ is the vector
of decision variables, $f_{i}:\RR^{n} \rightarrow {\RR}$, $i=1,\dots,k$ 
are the objective functions and $g_i,h_j:{\RR}^{n} \rightarrow {\RR}$,  $i=1,\dots,m$,
$j=1,\dots,p$ are the constraint functions of the problem. \\


\noindent {\bf Definition 1.} Given two vectors $\vec{x},\vec{y} \in \RR^k$,
we say that 
$\vec{x} \leq \vec{y}$ if $x_i \leq y_i$ for $i=1,...,k$, 
and that $\vec{x}$ {\bf dominates} $\vec{y}$ (denoted by $\vec{x}\prec\vec{y}$)
if $\vec{x}\leq \vec{y}$ and $\vec{x} \neq \vec{y}$.\\

\noindent {\bf Definition 2.} We say that a vector of decision variables 
$\vec{x}\in \mathcal{X}\subset \RR^n$ is {\bf nondominated} with respect
to  $\mathcal{X}$, if there does not exist 
another $\vec{x}'\in \mathcal{X}$ such that $\vec{f}(\vec{x}')\prec\vec{f}(\vec{x})$.\\

\noindent {\bf Definition 3.} We say that a vector of decision variables 
$\vec{x}^{*} \in \mathcal{F}\subset \RR^n$ ($\mathcal{F}$ is the feasible region)
is {\bf Pareto-optimal} if it is nondominated with respect to $\mathcal{F}$.\\

\noindent {\bf Definition 4.} The {\bf Pareto Optimal Set} $\mathcal{P^*}$ is defined by:

\[\mathcal{P^*}=\{\vec{x}\in \mathcal{F} |\vec{x}\textnormal{ is Pareto-optimal}\}\]

\noindent {\bf Definition 5.} The {\bf Pareto Front} $\mathcal{PF^*}$ is defined by:

\[\mathcal{PF^*}=\{ \vec{f}(\vec{x}) \in \RR^k |\vec{x}\in \mathcal{P^*}\}\]

\noindent When solving multi-objective optimization problems (MOPs),
the aim is to obtain the Pareto optimal set from the set $\mathcal{F}$.
Thus, given a MOP, the goal of a Multi-Objective Evolutionary Algorithm (MOEA) is 
to produce a good approximation of its Pareto front.
One of the most widely used MOEAs for problems having only two or three objectives is the Nondominated Sorting Genetic Algorithm-II  ({\it NSGA-II}) \cite{NSGA-II}. This MOEA 
solves a MOP using nondominated sorting and a crowding-comparison operator that acts as its density estimator.


\section{Approach and Algorithm}
We propose a nested-evolutionary algorithm for generating universal unrestricted adversarial examples in a black-box scenario.  Given a sequence of image filters as input, the algorithm returns the best image-agnostic filter configuration which, applied to the images from the dataset, greatly increases the classification error of the target model.

The method consists of two evolutionary nested algorithms: the outer algorithm is in charge of finding the sequence of filters to use, while the inner algorithm has to choose the parameter values.

The population is composed by sequences of parameterized filters that are applicable to images and transform them in possibly malign images.

Given a set $S = \{ f_1, f_2, \cdots f_m \}$ of $m$ image filters, the outer algorithm genotype (with length $l$) is encoded as a list of integers representing the corresponding filters in $S$. Similarly, the inner algorithm genotype is represented by a list containing the parameters used for each selected filter. 

The associated phenotype, applied to a set of images, generates the adversarial examples 
by applying the selected sequence of filters, with their corresponding optimized parameters, to legitimate images. 

\subsection{Outer Algorithm}
For the outer optimization  step we employ a genetic algorithm: a population of $N$ candidate solutions is iteratively evolved towards better solutions. In order to breed a new generation, population members are randomly selected and the crossover and mutation operations are performed. The quality of the candidates is evaluated based on their fitness values. 
\begin{description}
\item[Initial population:] it is generated by randomly selecting $l$ filters from the set $S$ of available filters and their parameters are initialized with default values equal to 1. 
\item[Crossover:] a standard one-point crossover is used to generate new off-springs from randomly selected members. Each child is guaranteed to inherit some genetic information from both parents, including the optimized parameters. 
\item[Mutation:] it is applied by substituting a filter with another one based on a mutation probability. The substituent filter is initialized with random parameter values. This way we also ensure a complete mutation of the parameters. 
\item[Selection:] at the end of each iteration, we choose the N best individuals from the set of 2N candidates (parents and offsprings) according to their fitness values. This process is repeated until the algorithm exhausts the allowed number of epochs.
\end{description}

\subsection{Inner Algorithm}
For the inner algorithm we propose and evaluate three different optimization strategies: a genetic algorithm (GA), a $(1,\lambda)$ evolutionary strategy (ES) and a random-based approach with tournament (in Algorithm \ref{GA_alg}, defined as follows: $optimizer_{O}$ , where $O$ $\in$ \emph{\{ 'GA', 'ES', 'Tournament' \}}).

The genetic algorithm of the inner optimization has the same structure and operators as the outer GA, except that its task is to evolve a population of lists of parameters for every individual from the outer algorithm.

Alternatively to GA, we propose to optimize the parameters by using $(1,\lambda)$ evolution strategy with $\lambda = 5$. ES iteratively updates a search distribution by following the natural gradient towards higher expected fitness. In our case, for each list of parameters we compute a batch of N samples by perturbing the original individual. A gradient towards a better solution is estimated using the fitness values of the N samples. This gradient is then used to update the original individual. The entire process is repeated until a stopping criterion is met. 

Finally, the random-based method is implemented as a 2-way competition. Given a solution inherited from the outer algorithm, a new individual is generated by randomly changing the parameters values of the original solution. The two candidates compete against each other in a tournament and the winner is passed on to the next generation.



\subsection{Evaluation}
The last part of our algorithm is about how the evaluation is performed. 
A candidate sequence of filters $\bar y$, and its own optimized parameters $\bar n$, is decoded as the phenotype $b$ which is evaluated by querying the target neural network.
We modeled the fitness function as a multi-objective problem which accounts for both the attack success rate as well as the detection mechanism bypassing rate. 
The goal is to give the attacker the ability to bypass detection mechanisms. We believe this to be a powerful feature of our method given that the field of adversarial machine learning lacks such approaches.

Let $F$ the target neural network,  $x_i$ the $i$-th image of the original dataset $X$ and X* the set of perturbed images $x^*_{i}$  obtained by applying the sequence of filters we want to evaluate to all the images in $X$, we define
\begin{itemize}
\item the \textit{Attack Success Rate} $ASR$ as
\begin{equation}
   ASR(X,X^*) =\frac{1}{n} \sum\limits_{i=0}^n {F(x_i) \neq F(x_i^*)}
   \end{equation}

where $n$ is the size of the dataset $X$ and $X^*$
\item the \textit{Detection Rate} $DR$ as
\begin{equation}
   DR(X^*) = \frac{1}{n} \sum\limits_{i=0}^n  D(x^*_i) 
\end{equation}
where $D$ is the chosen detector (e.g. feature squeezing), which returns $1$ if the image is detected as an attack, $0$ otherwise.
\end{itemize}
Accordingly, we can define multi-objective problem of our interest as
\begin{equation}
 minimize \: \mathcal{F}(X,X^*) = \{ 1.0-ASR(X,X^*), DR(X, X^*)\}
\end{equation}
which is managed by means of the {\it non\_dominated\_sorting} and {\it crowding\_distance} procedures of the NSGA-II technique \cite{NSGA-II}.
The general structure of the proposed algorithm is illustrated in Algorithm \ref{GA_alg}.

\begin{algorithm}[h]
\SetAlgoLined
\textbf{Input}: Dataset $D$, population size $N$, 
epochs $E$\\
 Extract the $K$ batches $B_1, \dots, B_K$ from $D$ \;
 Initialize population $P$ of $N$ individuals\;
 Evaluate each individual of $P$ by the two fitness $ASR$ and $DR$\;
  \For{$e=0$ \KwTo $E$}{ 
    \For{$i=1$ \KwTo $K$}{ 
        Offsprings = $\{ \emptyset \}$ \;
        \For{$i=1$ \KwTo $ N$}{
             Select randomly $parent_1$,$parent_2$ from $P$ \;
             $\overline p_1 \gets encode_{1}$($parent_1$) \;
             $\overline p_2 \gets encode_{1}$($parent_2$) \;
             $y_i$ = crossover($\overline p_1$,$\overline p_2$)\;
             $\overline{y_i}$ = mutation($y_i$) \;
             $ n_i \gets encode_{2}$($\overline y_i$) \;
             $\overline n_i$ = optimizer$_{O}(n_i)$ \;
             Offsprings $\gets$ ($\overline y_i$, $\overline n_i$) \;
        } 
        \ForEach{($\overline y_i, \: \overline n_i) \: \in \: \textup{Offsprings}$}{
             $b \gets$ decode ($\overline y_i$, $\overline n_i$) \;
             Evaluate the  fitness $ASR$ and $DR$ on batch $B_i$ \;
        }
        $P$ =  selection($P$, \textup{Offsprings}) \;
    } 
 } 
 
\textbf{return}: best image-agnostic filter configuration\;
 
 \caption{General structure of the nested evolutionary algorithm for generating adversarial examples}
 \label{GA_alg}
\end{algorithm}

\section{Experiments and Discussion}
\subsection{Experimental setup}
We evaluate the proposed method by attacking the convolutional neural network proposed by Papernot at al. in \cite{Papernot2016DistillationAA}  and used also in  \cite{Carlini2017TowardsET} to prove the effectiveness of their attack. The model is composed of a series of 2 convolutional layers having 64 3x3 filters paired with ReLU activation function and a max-pooling layer, 2 convolutional layers with 128 3x3 filters with ReLU followed by a another max-pooling layer, 2 fully connected layers with ReLU and a softmax layer used for the final classification. This network was trained using the CIFAR-10 dataset which is a very popular benchmark image dataset consisting in 50000 training and 10000 testing colour images with a resolution of 32x32, belonging to 10 different classes. Dropout was used in order to prevent overfitting, and momentum and parameter decay were employed to guarantee model convergence. 


For these preliminary experiments, we choose to adopt the {\it Feature Squeezing} detection method \cite{Xu2018FeatureSD} as detection method in the fitness function used during the optimization process since it is one of the most popular and low-cost techniques that has been proven to achieve high detection rates (over 85\% for CIFAR-10 and Imagenet dataset) against different famous state-of-the-art 
attacks. %



The hyperparameters default values used to conduct the experiments were fixed as follows, where not differently specified: number of filters = 5, mutation probability = 0.5, batch size = 100, population size = 10 for the outer algorithm, epochs = 3. For the inner algorithms we set the population size equal to 5 and the number of generations was fixed to 3.  To perform the detection we used the combination of features squeezers reported in \cite{Xu2018FeatureSD} to work best for CIFAR-10 images: reduction to 5-bit depth, a local median smoothing and a non-local mean smoothing, and threshold to find the illegitimate images set to 1.7547 
\footnote{ \url{https://github.com/mzweilin/EvadeML-Zoo/blob/master/Reproduce_FeatureSqueezing.md}}.

\noindent\textbf{Dataset}

We used the CIFAR-10 testing set for training our algorithm and evaluating its effectiveness. The set was divided in two subsets: the first  200 images were used for the filter configuration optimization process and the remaining 9800 images were used for testing the adversarial attack. The optimization subset of images was chosen relatively small in order to measure the power of the universal attack.\\

\noindent\textbf{Selection of the training epochs, number of filters and parameters range}

Several experiments were carried out in order to estimate the best trade-off between the performance of the proposed method and computation time. We tested all three inner optimization algorithms (GA, ES and Tournament) with the default parameters configuration except for the number of epochs which was set to 10. We analyzed their attack success rate (ASR), feature squeezing detection rate (DR) and computation time. 
We observed that they all had similar performance-time behaviour. We decided to stick to 3 epochs since it was producing good results while keeping the computational time fairly low. Figure \ref{fig1} illustrates the attack and detection rate curve with respect to the number of epochs with ES inner optimizer.

{\renewcommand{\arraystretch}{1.0}
\begin{figure}[h]
\centering
\includegraphics[width=0.9\textwidth]{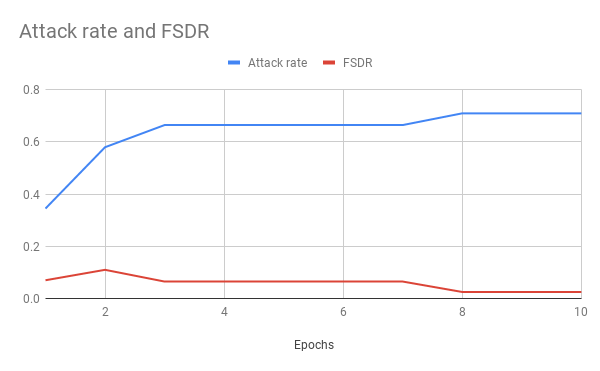}
\caption{Attack rate and FSDR wrt Epochs with ES optimizer.} \label{fig1}
\end{figure}
}
Moreover, we also wanted to investigate the importance of choosing different numbers of filters for creating the adversarial configuration. The minimum filters selection was set to 3 while the maximum is the cardinality of set $S$ of available filters. We adopted the policy of no-repeating filters, meaning that a filter can be picked only once inside a certain configuration.
We calculated the attack rate of our algorithm by using all three inner optimization methods. Table \ref{risultati_numero_filtri} shows that using 5 filters has the best outcome in terms of attack success rate. 
{\renewcommand{\arraystretch}{1.0}
\begin{table}[h]
\caption{Evaluation of attack success rate(ASR) with respect to the number of filters. }
\centering
\begin{tabular}{|c|c|c|}
\hline
Optimizer & Number of filters & ASR \% \\
\hline
ES  &3&   46.5 \\ 
\hline
ES  &4 &   43.5 \\
\hline
\bf ES  &\bf 5 & \bf 70  \\
\hline
GA  &3 & 58.5\\ 
\hline
GA  &4 &  52 \\
\hline
\bf GA  &\bf 5 & \bf68.5 \\
\hline
Tournament & 3 & 41.5 \\
\hline
Tournament  &4 &  45.5 \\
\hline
\textbf{Tournament} & \textbf{5} & \textbf{61}\\
\hline

\end{tabular}
\label{risultati_numero_filtri}
\end{table}
}

In our implementation filters can be applied using different features parameters similar to how Instagram allows users to control the effect of filters by manually adjusting their intensities within  a certain range. 
The parameters of each filter can vary between a fixed range of values. The minimum and maximum values of each interval were found by performing a quality analysis on the modified images with the above mentioned filters and diverse parameters values. 
This analysis allowed to restrict the search space in order to further reduce the training time.
In order to evaluate the universality of our attack we applied the optimized filter configuration to each image in the testing set and computed the detection rate defined as follows:
\begin{equation}
    FSDR = \frac{\sum \limits_{i=0}^m D(\widehat{x_i})}{|\widehat{X}|},\qquad \widehat{x} \in \widehat{X}
\end{equation}
 where $D$ corresponds to the features squeezing detector which returns $1$ if the image is identified as illegitimate and $0$ otherwise, $\widehat{X}$ represents the set of successful adversarial examples,  
and $m$ = $|\widehat{X}|$ is the cardinality of $\widehat{X}$. 

In Table \ref{risultati_best} we report the attack success rate and the detection rate for both training and testing subsets with the default hyperparameters values, which were found to work best. 

{\renewcommand{\arraystretch}{1.0}
\begin{table}[h]
\caption{ Attack success rate (ASR) and Feature Squeezing Detection Rate (FSDR) with different optimizers on Carlini CNN and CIFAR-10 training and testing subsets, epochs = 3, number of filters = 5. }
\centering
\begin{tabular}{|c|c|c|c|c|}
\hline
Optimizer & ASR \% train set & FSDR \% train set & ASR \% test set & FSDR \% test set\\
\hline
ES & 70 &  2.1 & 63.7 & 3.5 \\
\hline
GA & 68.5 &   2.9 & 63.8 & 3.4\\
\hline
Tournament & 61 &  5.7& 56.3& 4.5\\
\hline

\end{tabular}
\label{risultati_best}
\end{table}
}

First of all, from these results, we can note that, even if the attack success rate is lower than the ones obtained by other methods in literature (also greater than 90\% in some cases), these values should be fairly compared to the ones obtained by the other methods excluding the attacks that would be blocked by a defense mechanism. Considering that Xu et al. evaluated Feature Squeezing method with respect to 11 different attacks on three different datasets and reported for CIFAR-10 an overall detection rate of 84.5\%  \cite{Xu2018FeatureSD},
our attack is very effective 
because among the successful adversarial images just very few attempts will be blocked by the defense mechanism.
Moreover, we can observe a very good generalization ability of the model: when the attack model generated by our method is applied to the test set, we lose less than 10\% for ASR maintaining a very low detection rate when a defense mechanism based on Feature Squeezing method is applied.

\subsection{Generated Images}
Table \ref{risultati_best_images} shows some successful adversarial examples generated by applying  the filter configurations with their respective optimized parameters found by the proposed algorithm on the unseen images from the testing subset. 

For each adversarial example we attached the original image and we also indicate the classification labels before and after the modification. It is very interesting to note that the solutions found by our method, i.e. the  applied perturbations, are very uniform across the image and no unnatural patterns or high-frequency areas can be noticed.

{\renewcommand{\arraystretch}{1.0}
\begin{table}[h!]
\caption{Successful adversarial attacks on CIFAR-10 testing subset. 
On the left: original image; On the right: successful adversarial example.}
\centering
\begin{tabular}{|c|c|}
\hline
Optimizer & Successful adversarial examples on the testing set \\
\hline
 &  \includegraphics[scale=1]{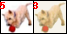} 
\includegraphics[scale=1]{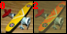} 
\includegraphics[scale=1]{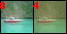} 
\includegraphics[scale=1]{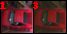} \\

 ES & \includegraphics[scale=1]{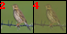} 
 \includegraphics[scale=1]{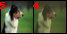} 
 \includegraphics[scale=1]{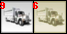}
 \includegraphics[scale=1]{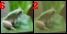} \\

 & \includegraphics[scale=1]{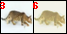} 
\includegraphics[scale=1]{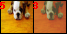} 
\includegraphics[scale=1]{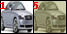} 
\includegraphics[scale=1]{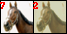}  \\
 
  & \includegraphics[scale=1]{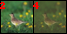} 
\includegraphics[scale=1]{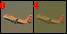} 
\includegraphics[scale=1]{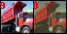} 
\includegraphics[scale=1]{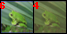}\\
 
\hline
 &  \includegraphics[scale=1]{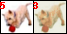} 
\includegraphics[scale=1]{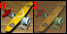} 
\includegraphics[scale=1]{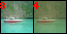}
\includegraphics[scale=1]{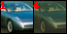} \\

GA &  \includegraphics[scale=1]{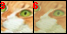} 
\includegraphics[scale=1]{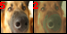} 
\includegraphics[scale=1]{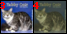}
\includegraphics[scale=1]{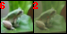} \\

&  \includegraphics[scale=1]{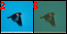} 
\includegraphics[scale=1]{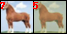} 
\includegraphics[scale=1]{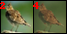} 
\includegraphics[scale=1]{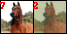} \\

&  \includegraphics[scale=1]{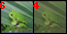}
\includegraphics[scale=1]{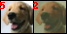}
\includegraphics[scale=1]{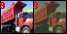}
\includegraphics[scale=1]{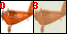} \\
\hline

 &  \includegraphics[scale=1]{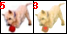} 
\includegraphics[scale=1]{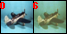} 
\includegraphics[scale=1]{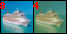}
\includegraphics[scale=1]{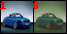} \\

Tournament &  \includegraphics[scale=1]{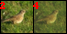} 
\includegraphics[scale=1]{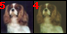} 
\includegraphics[scale=1]{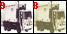}
\includegraphics[scale=1]{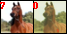} \\

&  \includegraphics[scale=1]{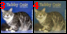} 
\includegraphics[scale=1]{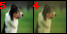} 
\includegraphics[scale=1]{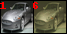} 
\includegraphics[scale=1]{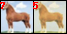} 
\\

&  \includegraphics[scale=1]{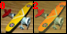} 
\includegraphics[scale=1]{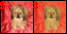} 
\includegraphics[scale=1]{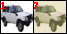} 
\includegraphics[scale=1]{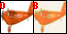}  
 \\
\hline

Label names &     airplane : 0,   automobile : 1, bird : 2,  cat : 3, deer : 4,   \\

 &       dog : 5,    frog : 6,   horse : 7,   ship : 8,   truck : 9 \\
\hline
\end{tabular}
\label{risultati_best_images}
\end{table}
}



\section{Conclusions and Future Works}

The experimental results show that the multi-objective method with detection feedback is able to produce successful adversarial examples while keeping the detection rate low. Even though the attack success rate is lower with respect to other state-of-the-art methods (restricted \& unrestricted) we have the advantage of not being caught by detection methods.
This indicates the potential of the proposed attack whose goal is not only to force the classifier to mispredict but also to evade possible defenses. 

Nonetheless, there is a wide room for improvement. We intend to continue this study since the topic of multi-objective evolutionary attacks opens up an interesting research direction. We also plan to run more experiments on more complex models and to test the universality of our attacks across multiple neural networks architectures. Moreover, considering the availability of multi-objective evolutionary methods like MOEA/D \cite{MOEAD} able to take into account more than two objective functions, we want to further improve our algorithm in order to include in the fitness function the three components: attack success rate, detection rate and image control.
 Differently from all the other approaches, our idea is to manage the image perturbations and control it by using 
  no-reference image quality assessment like  NIMA \cite{Talebi2018NIMANI} or
  Frechet Inception Distance (FID) \cite{fid_NIPS2017_8a1d6947} instead of the $L_p$-norms. 


%
%
%
 \bibliographystyle{LLNCS/splncs04}
\bibliography{biblio}

\end{document}